\documentclass[conference]{IEEEtran}
\IEEEoverridecommandlockouts

\usepackage{amsmath,amssymb,amsfonts}
\usepackage{algorithmic}
\usepackage{graphicx}
\usepackage{textcomp}
\usepackage{xcolor}

\usepackage{caption}
\usepackage{subcaption}
\usepackage{hyperref}

\newcommand{\orcid}{\includegraphics[scale=0.07]{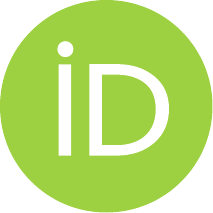} \hspace{1mm}}

\def\titlethis{Gap and Overlap Detection in Automated Fiber Placement}
\def\titleshortthis{Gap and Overlap Detection in AFP}
\def\keywordsthis{automated fiber placement, quality inspection, computer vision, image processing, edge detection}

\hypersetup{
pdftitle = \titleshortthis,
pdfsubject = \titlethis
pdfauthor = {Assef Ghamisi, Homayoun Najjaran},
pdfkeywords = {\keywordsthis},
pdflang = {en-US},
colorlinks = true,
linkcolor = blue,
urlcolor = black,
citecolor = purple,
linktocpage = true,
}

\begin{document}

\title{\titlethis}

\author{\IEEEauthorblockN{1\textsuperscript{st} \href{https://orcid.org/0000-0003-2961-5992}{Assef Ghamisi \orcid}}
\IEEEauthorblockA{\textit{Electrical and Computer Engineering} \\
\textit{University of Victoria}\\
Victoria, Canada \\
assefghamisi@uvic.ca}
\and
\IEEEauthorblockN{2\textsuperscript{nd} \href{https://orcid.org/0000-0002-3550-225X}{Homayoun Najjaran \orcid}}
\IEEEauthorblockA{\textit{Mechanical Engineering} \\
\textit{University of Victoria}\\
Victoria, Canada \\
najjaran@uvic.ca}
}

\maketitle

\begin{abstract}
    The identification and correction of manufacturing defects, particularly gaps and overlaps, are crucial for ensuring high-quality composite parts produced through Automated Fiber Placement (AFP). These imperfections are the most commonly observed issues that can significantly impact the overall quality of the composite parts. Manual inspection is both time-consuming and labor-intensive, making it an inefficient approach. To overcome this challenge, the implementation of an automated defect detection system serves as the optimal solution. In this paper, we introduce a novel method that uses an Optical Coherence Tomography (OCT) sensor and computer vision techniques to detect and locate gaps and overlaps in composite parts. Our approach involves generating a depth map image of the composite surface that highlights the elevation of composite tapes (or tows) on the surface. By detecting the boundaries of each tow, our algorithm can compare consecutive tows and identify gaps or overlaps that may exist between them. Any gaps or overlaps exceeding a predefined tolerance threshold are considered manufacturing defects. To evaluate the performance of our approach, we compare the detected defects with the ground truth annotated by experts. The results demonstrate a high level of accuracy and efficiency in gap and overlap segmentation.
\end{abstract}
\section{Introduction}  \label{sec:intro}

Automated fiber placement (AFP) is an advanced manufacturing technique used to produce composite materials. It involves the precise robotic placement of continuous fiber tapes onto a surface to create strong and lightweight structures. The composite parts manufactured using AFP are used in the aerospace and automotive industries \cite{brasington_automated_2021, zhang_review_2020}. However, like any manufacturing process, AFP is not immune to defects, significantly impacting the mechanical properties of the composite part \cite{palardy-sim_next_2019, bockl_effects_2023}. For example, the presence of systematic defects may decrease the strength of the composite structure \cite{woigk_experimental_2018}. Therefore, inspection plays a crucial role in AFP to detect defects and ensure defect-free composite parts.

\begin{figure}[t]
\begin{center}
   \includegraphics[width=0.8\linewidth]{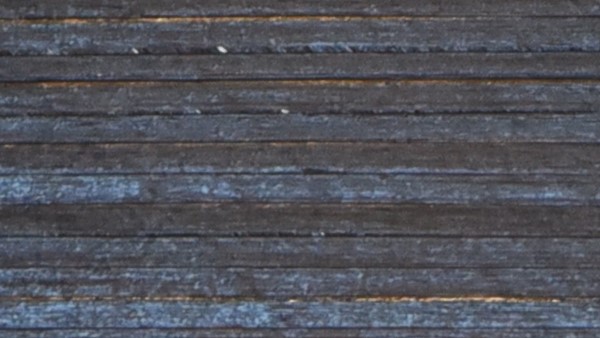}
\end{center}
   \caption{A photo of the AFP-manufacture structure shows gaps and overlaps between the composite tows.}
\label{fig:real}
\end{figure}

Gaps and overlaps are the most frequent defects in AFP \cite{heinecke_manufacturing-induced_2019,harik_automated_2018, oromiehie_automated_2019}. A gap happens when two neighboring tows are misaligned in a way that there is some space between them. On the other hand, an overlap occurs when the two adjacent tows are positioned such that they partially cover each other \cite{harik_automated_2018}. A close photograph of the AFP surface is provided in Fig. \ref{fig:real}, showing gaps in the lighter and overlaps in the darker areas. The presence of gaps and overlaps can negatively impact the mechanical properties of the final part. They may results in uneven distribution of thickness \cite{oromiehie_automated_2019}, and strength reduction \cite{fayazbakhsh_defect_2013} in the laminate. Also, they can form regions within the composite part that are excessively rich in either resin or fiber \cite{lan_influence_2016}.

Manual inspection is often used to detect defects like gaps and overlaps, but it is not always accurate or time-efficient \cite{meister_review_2021}. In some cases, manual inspection can even take longer than AFP deposition itself \cite{heinecke_manufacturing-induced_2019}. Consequently, there is ongoing research focused on automating defect detection in AFP.

Many non-destructive testing methods have been used to evaluate composite structures \cite{gholizadeh_review_2016}. Some of these methods utilize profilometry to capture surface information from composite parts and present it in the form of images. Various imaging tools employ different sensors, such as optical cameras \cite{shadmehri_laser-vision_2015}, thermal cameras \cite{juarez_situ_2021, denkena_thermographic_2016}, and laser scanners \cite{cemenska_automated_2015, maass_progress_2015}. Researchers at the National Research Canada (NRC) have developed a precise imaging tool using an Optical Coherence Tomography (OCT) sensor \cite{roy_automated_2019}. In this paper, we present our inspection tool based on OCT sensor data. However, our method can be scaled to other AFP imaging techniques with some modifications.

To utilize AFP imaging for automatic defect detection, recent research has explored both machine learning and rule-based approaches. ML-based methods usually take advantage of Convolutional Neural Networks (CNNs) to detect defects \cite{sacco_machine_2020, schmidt_deep_2019, zhang_research_2022, ghamisi_anomaly_2023}. These methods are general but they typically require large amounts of data, particularly defective samples, which may not be widely available. On the other hand, classical computer vision techniques are not data-driven, and they work based on rules that can be customized to address specific challenges in the defect detection process.

In this paper, we propose a novel approach to gap and overlap detection in AFP that does not require training any machine learning model and thus solving the data scarcity problem in AFP. Our method leverages the understanding of the format of the composite tapes (tows) to create a customized tool that can segment gaps and overlaps in composite parts. We evaluate our approach on samples of composite profilometry images and compare its performance with existing methods. Our experimental results demonstrate that our approach can achieve high accuracy in gap and overlap segmentation, making it a promising tool for improving the efficiency and quality of the AFP process.

Section \ref{sec:method} provides a detailed explanation of the algorithm, including its implementation details. In Section \ref{sec:results}, we present a comprehensive analysis of the gap and overlap segmentation tool, showcasing the step-by-step results and evaluating the impact of various algorithm components. Lastly, Section \ref{sec:conclusions} concludes this work and offers valuable insights for future research endeavors.

Our main contribution is to present a novel algorithm for the detection and demonstration of gaps and overlaps in the process of automated fiber placement. We also implemented and tested the algorithm to prove the feasibility of this approach.

\section{Methodology} \label{sec:method}

\begin{figure}[t]
\begin{center}
   \includegraphics[width=0.9\linewidth]{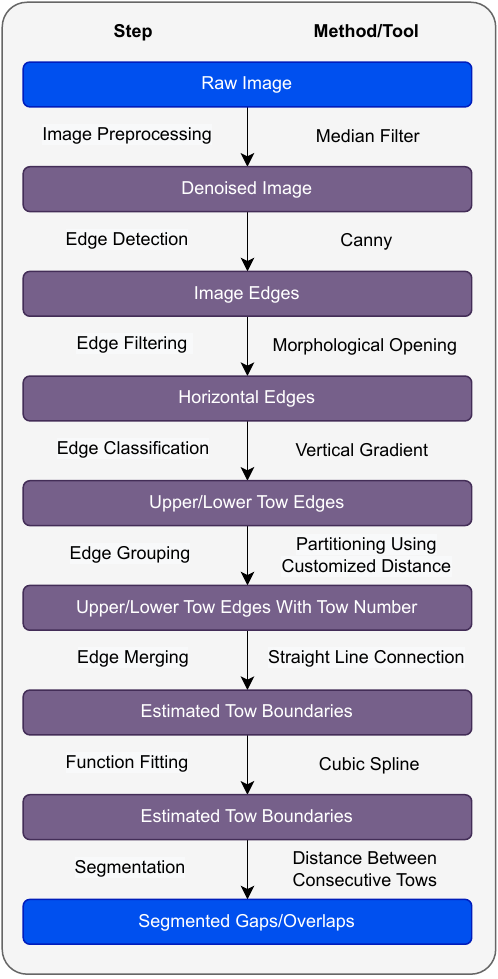}
\end{center}
   \caption{A summary of the proposed gap and overlap segmentation method.}
\label{fig:steps}
\end{figure}

This section explores the procedures involved in our gap and overlap segmentation algorithm. Initially, we employ capture scans from composite structures and apply fundamental image processing techniques to enhance the input image, eliminate unwanted noise. Then, we extract the edges on the image and filter only the horizontal edges that correspond to the upper and lower boundaries of the tows. Subsequently, we proceed to group and merge the extracted edges to create continuous horizontal edges for the tows. Following that, we apply a curve-fitting approach to estimate and represent the boundaries of the tows using the merged edges. Lastly, we utilize the tow boundaries to segment any gaps and overlaps in the image. Fig. \ref{fig:steps} summarizes our methodology by presenting each step, the used method, and their input and output.

\subsection{Image Acquisition and Preprocessing}

\begin{figure}[t]
\begin{center}
   \includegraphics[width=0.9\linewidth]{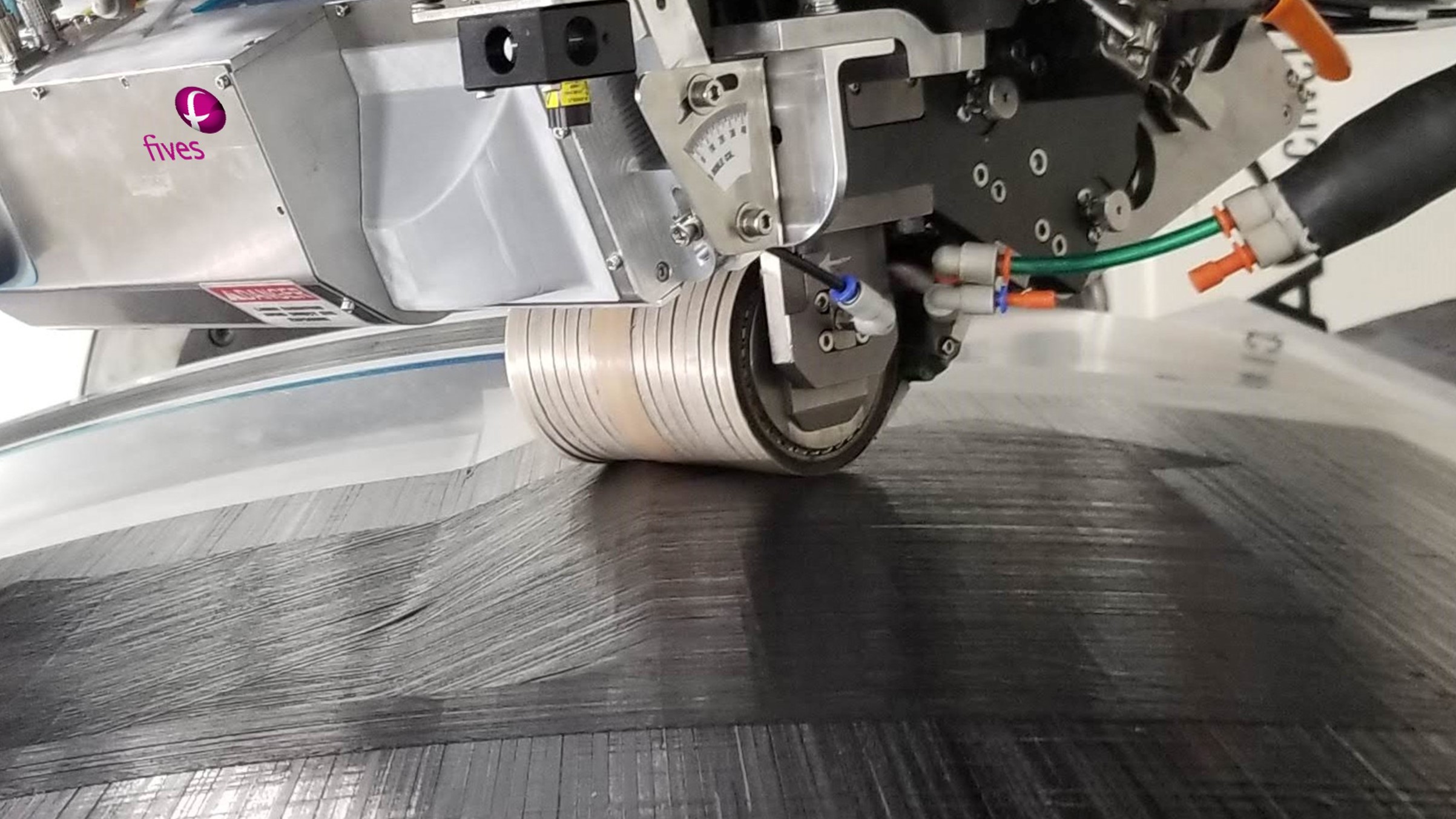}
\end{center}
   \caption{AFP machine equipped with an OCT sensor.}
\label{fig:afp_machine}
\end{figure}

\begin{figure}[t]
\begin{center}
   \includegraphics[width=0.7\linewidth]{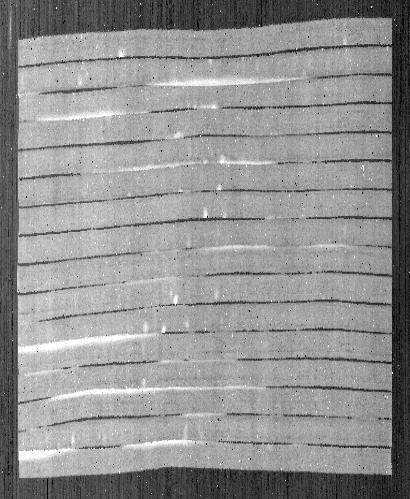}
\end{center}
   \caption{A specimen scan of the composite surface generated by OCT sensor.}
\label{fig:original}
\end{figure}

We utilize an OCT sensor mounted on top of the automated fiber placement (AFP) head to produce scans of the composite structures, as shown in Fig. \ref{fig:afp_machine}. Fig. \ref{fig:original} shows a sample of scans generated. The sensor scans are presented as a grayscale image, where the brightness of each pixel corresponds to the height of the composite part. Additionally, the horizontal and vertical position of each pixel in the image corresponds to the Cartesian position of the corresponding point on the surface of the composite part \cite{rivard_enabling_2020}.

The raw images contain black and white dots (also known as salt and pepper noise), which may negatively impact the defect detection process. To eliminate these dots, we employ a median filter \cite{alqadi_salt_2018} with a kernel size of $N\times N$. For optimal results, we opt for a moderate filter size of ($N=3$) to effectively remove the majority of these dots while preserving the essential visual details within the image.

\subsection{Edge Detection and Filtering}

\begin{figure}[t]
\begin{center}
   \includegraphics[width=0.7\linewidth]{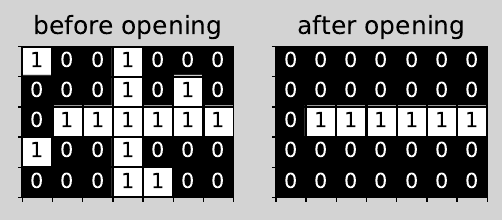}
\end{center}
   \caption{A morphological opening operation with a horizontal line structuring element only preserves the horizontal edge pieces.}
\label{fig:opening}
\end{figure}

Edges in grayscale images are sharp transitions in intensity that mark the boundaries between different regions within the image. In AFP images, some edges may flag the boundaries of the tows which are useful for gap and overlap detection. To extract the edges of the image, we employ canny edge detection with a proper Gaussian filter size \cite{canny_computational_1986}. While this process captures all edges, some of them may not correspond to the upper or lower boundaries of the tows. Since our objective is to isolate only the upper and lower boundaries, it becomes necessary to eliminate any additional information or noise present in these edges. To accomplish this, we utilize an opening morphological operation \cite{haralick_image_1987}. To allow horizontal edge pieces, we use a horizontal line with a length of 5 pixels as the structuring element for the opening operation. Fig. \ref{fig:opening} shows the impact of the opening operation on filtering out noise and non-horizontal edges.

\begin{figure}
     \centering
     \begin{subfigure}[b]{0.8\linewidth}
         \centering
         \includegraphics[width=\textwidth]{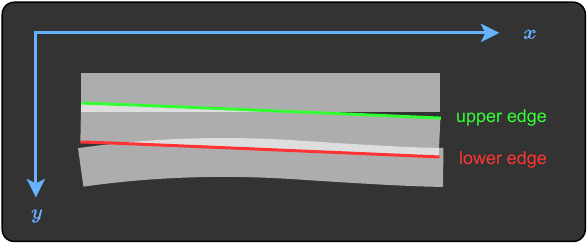}
         \caption{Upper and lower boundaries of a tow.}
         \label{fig:upper_lower}
     \end{subfigure}
     \hfill
     \begin{subfigure}[b]{0.8\linewidth}
         \centering
         \includegraphics[width=\textwidth]{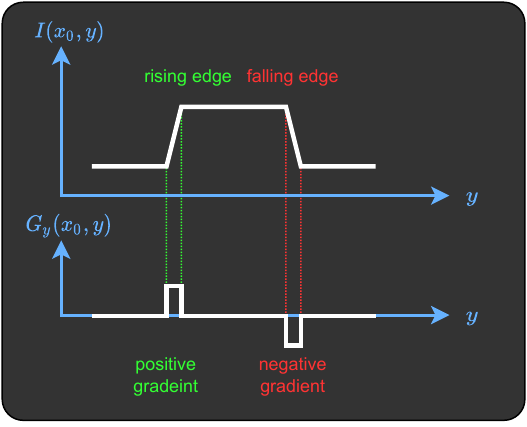}
         \caption{Presence of a tow causes vertical changes in image intensity and gradient sign.}
         \label{fig:image_gradient}
     \end{subfigure}
        \caption{Distinction between upper and lower tow edges.}
        \label{fig:edges_method}
\end{figure}

After extracting all the horizontal edges from the image, the next step is to classify these edge pieces into two distinct categories depending on whether they correspond to the upper or lower boundaries of the tows (Fig. \ref{fig:upper_lower}). The challenge in this task is that the lower edge of the upper tow and the upper edge of the lower tow are often very close to each other, and sometimes they overlap. To tackle this problem, we rely on the sign of the vertical gradient of the image as the classification criteria. To begin with, a Sobel operator \cite{sobel_isotropic_2014} formulated as in Equation \ref{eq:gradient} calculates the vertical gradient denoted as $G_y$. Here, $I$ represents the input image, and $*$ signifies a 2D convolution.

\begin{equation}
G_y=\left[\begin{array}{ccc}
+1 & +2 & +1 \\
0 & 0 & 0 \\
-1 & -2 & -1
\end{array}\right] * I
\label{eq:gradient}
\end{equation}

Considering the up-to-down direction for the vertical axis (rows of the image) in Fig. \ref{fig:image_gradient}, a rising edge in the pixel brightness values corresponds to the upper edges of a tow. As a result, positive values of $G_y$ indicate the presence of upper boundaries. Conversely, negative values signify lower boundaries of the tows. Thus, by examining the values of the vertical gradient, we can effectively classify the extracted horizontal edges as either upper (positive) edges or lower (negative) edges.

\subsection{Merging the Edges}

The identified horizontal edges are discontinuous, and as a result, they do not show the upper and lower boundaries of the tows. To address this, it is necessary to merge the edges associated with the same tow, effectively creating continuous boundaries.

\begin{figure}[t]
\begin{center}
   \includegraphics[width=0.6\linewidth]{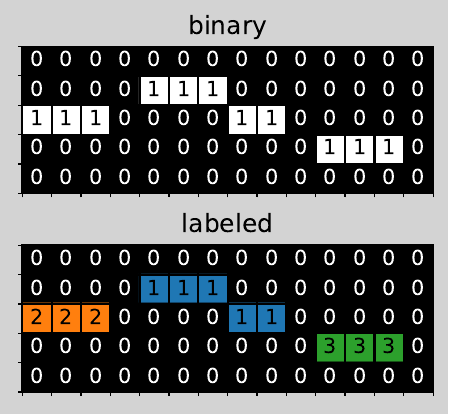}
\end{center}
   \caption{Connected components labeling of edge regions.}
\label{fig:labeling}
\end{figure}

First, we make use of a computer vision method called Connected Component Labeling \cite{he_connected-component_2017} to identify and distinguish each connected component (region) of the binary image as illustrated in Fig. \ref{fig:labeling}. In this project, each region is a piece of the horizontal edge of the tow that is not connected to other edges.

We need to accurately group the edge regions that belong to the same tow to prevent a wrong connection between two regions in different tows. To achieve edge groping we define an equivalency relation ($R$) provides in \ref{eq:relation}. The regions $a$ and $b$ are considered to be in the same class (tow) if their distance ($d$) is lower than the threshold value $d_{Th}$.

\begin{equation}
\begin{aligned}
R = \{ (a, b) : d(a,b)<d_{Th} \}
\end{aligned}
\label{eq:relation}
\end{equation}

More importantly, they need to be vertically close to each other.
Since vertical distance is more effective than horizontal distance, we use a customized distance function that considers both horizontal and vertical distances of the regions with more penalty on vertical distance. This distance is formulated as equation \ref{eq:distance}. In this equation, $d$ is the customized distance, $\Delta x$ and $\Delta y$ are horizontal and vertical distances, respectively. Also, $a_1$ and $a_2$ are constant coefficients.

The customized distance between each pair of regions is calculated. If two regions are close, they are considered in the same equivalence class. In this way, we group all the regions into some classes, each class represents a unique tow in the image.

\begin{equation}
\begin{aligned}
d(a,b) = \alpha_x \lvert x_b-x_a \rvert + \alpha_b \lvert y_b-y_a \rvert \quad \textrm{,} \quad \alpha_x < \alpha_y
\end{aligned}
\label{eq:distance}
\end{equation}

\begin{figure}[t]
\begin{center}
   \includegraphics[width=1\linewidth]{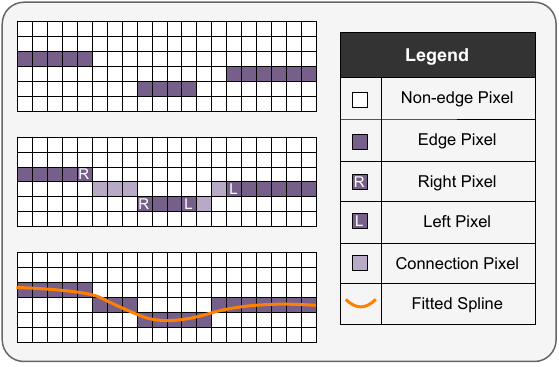}
\end{center}
   \caption{Connecting the edge regions within a tow.}
\label{fig:connectivity}
\end{figure}

After grouping the regions that belong to the same tow, we can merge them. We connect region pairs that are horizontally next to each other. To achieve this, we use a straight line to connect the rightmost pixel of the left region to the leftmost pixel of the right region. The process of merging these edges to form cohesive boundaries of the tows is visually depicted in Fig. \ref{fig:connectivity}.

To obtain a clear representation of reconstructed tow boundaries, we transform the merged tow edges (pixels) into smooth curves (vectors) for visualization. This is accomplished by applying a univariate spline \cite{greiner_survey_1991} fitting to the merged edges of each tow. For each tow edge, we have a list of pixel positions that are strictly ascending in terms of horizontal positions. We find a function ($f$) that, given the horizontal positions of the edge pixel ($x$), can approximate the vertical position of the edge pixel ($y = f(x)$).

\subsection{Defect Segmentation}
The estimated boundaries of the tows can be processed to segment gaps and overlaps on the input image. When the upper boundary of the lower tow is higher than the lower boundary of the upper tow, it is possible that an overlap between the two tows happened. On the other hand, if the former is lower than the latter, a gap may exist between the tows corresponding to those curves. Based on these criteria the pixels in the input image can be classified into one of three classes neutral, gap, or overlap. 
Note that the segmented areas are identified as potential manufacturing defects. In fact, gaps and overlaps within the acceptable tolerance levels are not considered defects. Further investigations are necessary to determine whether these identified gaps and overlaps should be classified as defects or not.

\subsection{Performance Evaluation}
We assess our findings by comparing the segmentation results to ground truth. The ground truth is established by an operator who labels the images. This labeling process involves using a brush tool to mark all the pixels that belong to the gap and overlap categories. To evaluate the performance of the gap/overlap segmentation algorithm, we utilize Intersection over Union (IoU) as a metric for each of these categories. By calculating the IoU values for gaps and overlaps and averaging them, we obtain an overall measure of the algorithm's performance.

\section{Results and Discussion} \label{sec:results}

\begin{figure}[t]
\begin{center}
   \includegraphics[width=0.9\linewidth]{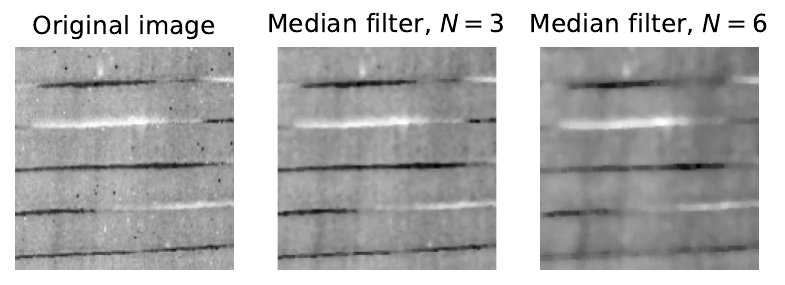}
\end{center}
   \caption{Applying median filters with different window sizes.}
\label{fig:median_filtered}
\end{figure}

In Fig. \ref{fig:median_filtered} the impact of median filters on the image is demonstrated. Using a large median filter ($N=6$) results in losing important features like the edges of the tows. Whereas, choosing an average median filter ($N=3$) reduces salt and pepper noise while preserving the edges of the tows.

\begin{figure}[t]
\begin{center}
   \includegraphics[width=0.9\linewidth]{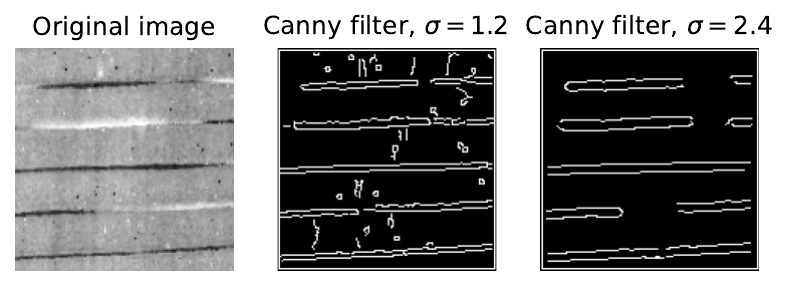}
\end{center}
   \caption{Canny edge detection with different Gaussian filter sizes.}
\label{fig:canny}
\end{figure}

Fig. \ref{fig:canny} shows how a canny edge detector extracts the edges on the image. By increasing the value standard deviation ($\sigma$) for the Gaussian filter, a clearer representation of tow edges is extracted but some edges are missed.
 
\begin{figure}[t]
\begin{center}
   \includegraphics[width=0.9\linewidth]{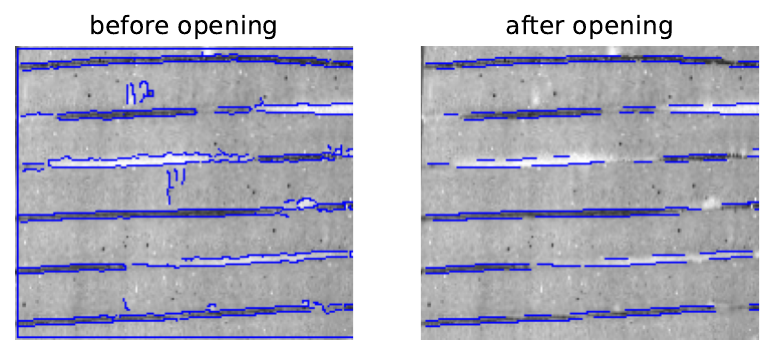}
\end{center}
   \caption{Edge detection before and after the opening operation.}
\label{fig:edges_opening}
\end{figure}

To filter out undesired edges, a morphological opening operator is used that only preserves the horizontal edges. The detected edges before and after applying the morphological operation are demonstrated in Fig. \ref{fig:edges_opening}. 

\begin{figure}[t]
\begin{center}
   \includegraphics[width=0.9\linewidth]{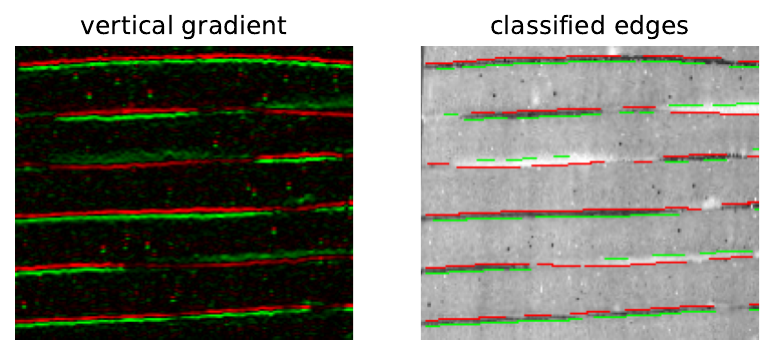}
\end{center}
   \caption{Positive and negative components of the horizontal gradient (left) are used to classify the edges into top and bottom edges of the tows (right).}
\label{fig:edge_classification}
\end{figure}

Fig. \ref{fig:edge_classification} illustrates the result of tow edge classification. The Sobel operator generates the horizontal gradient of the image. We use green and red colors to illustrate positive and negative gradient values. Also, The pixel brightness on the gradient image shows the normalized magnitude of the gradient. The gradient values are positive for the lower edges of the tows and negative for the upper edges of the tows.
Considering the sign of the horizontal gradient, the edges are classified into either upper or lower edges of the tows which are distinguished by green and red colors, respectively.

\begin{figure}[t]
\begin{center}
   \includegraphics[width=0.9\linewidth]{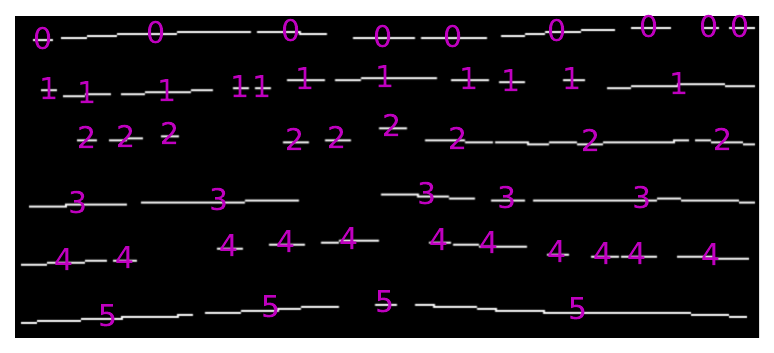}
\end{center}
   \caption{Grouping edge regions. Each region is labeled with the corresponding group number.}
\label{fig:edge_grouping}
\end{figure}

 Fig. \ref{fig:edge_grouping} demonstrates the result of edge grouping for the upper tow boundaries. We used a customized distance to partition the edge pieces into equivalency classes. Each edge is annotated with the detected equivalency class number. As expected, the edge pieces in the same tow are grouped into the same class.

\begin{figure}[t]
\begin{center}
   \includegraphics[width=0.9\linewidth]{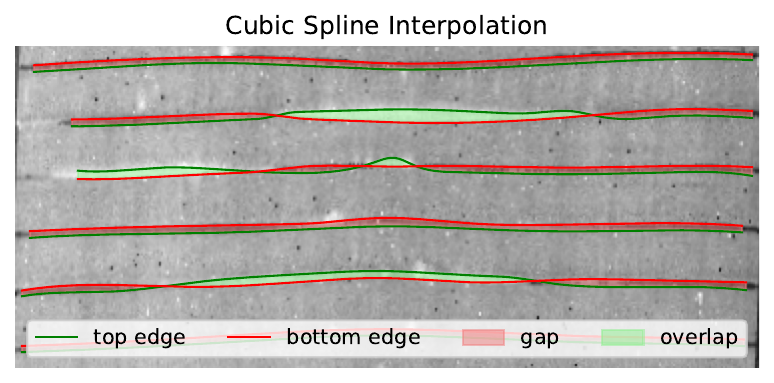}
\end{center}
   \caption{Reconstructed horizontal boundaries of tows reveal gaps and overlaps.}
\label{fig:reconstructed}
\end{figure}

Fig. \ref{fig:reconstructed} illustrates the result of horizontal edge reconstruction. The upper and lower edges of the tows are interpolated by a cubic spline and annotated on the original image. The areas between these edges are also annotated as gaps or overlaps. The software was unable to detect part of the edges on the left and right sides of the tows. This usually happens when the two consecutive tows are perfectly laid close to each other without any gaps or overlaps. In that case, we are not missing any gaps or overlaps.

\begin{figure}[t]
\begin{center}
   \includegraphics[width=0.9\linewidth]{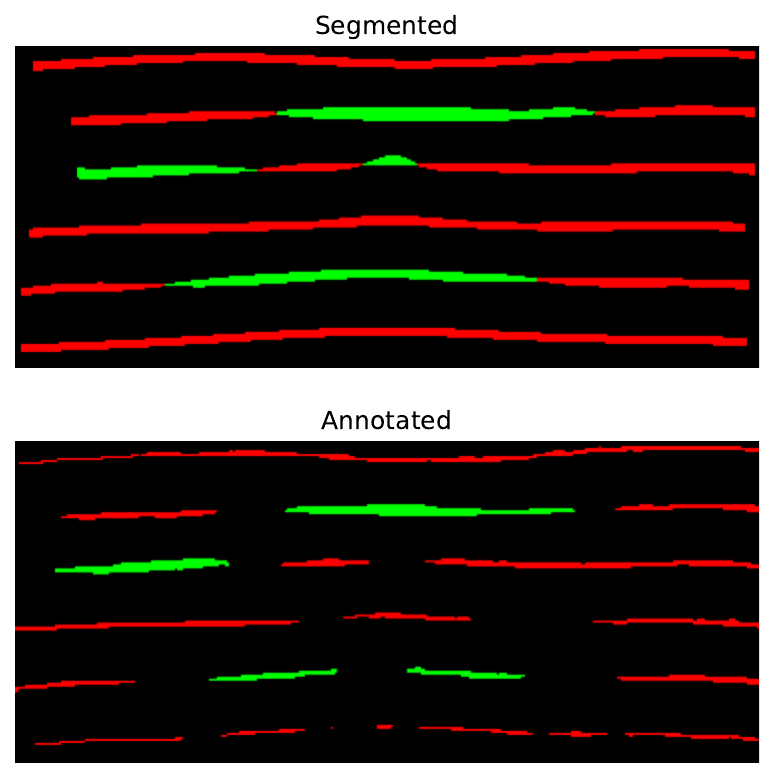}
\end{center}
   \caption{A comparison between gap and overlap: predicted segmentation (top) and annotated ground truth (bottom). Gaps are shown in red and overlaps in green.}
\label{fig:segmentation}
\end{figure}

A comparison between the segmented image and annotated ground truth is demonstrated in Fig. \ref{fig:segmentation}. The algorithm detects most of the gaps and overlaps areas which was the final goal of this work. The values of intersection over Union (IoU) are 0.381, and 0.494 for the gap and overlap classes, respectively. This is equal to an average IoU of 0.438 for both classes.

\section{Conclusions and Future Work} \label{sec:conclusions}

Our work presents a computer vision algorithm for detecting and segmenting gaps and overlaps in composite structures fabricated through automated fiber placement (AFP). Our software takes a depth scan of the composite part as input and generates a spline representation of individual gaps and overlaps, enabling operators to locate and identify them easily. We have also investigated the influence of various design parameters on the performance of the segmentation pipeline and evaluated it using the intersection over union metric, resulting in an average value of 0.438 for the gap and overlap classes.

In future steps, the detected boundaries of the tows in this work can serve as a starting point for identifying other types of manufacturing defects in AFP. Moreover, utilizing the extracted gaps and overlaps, an automatic system can be developed to classify them as defects or non-defects based on their shape and size. The tape-by-tape approach employed in this study, with appropriate modifications, can also be applied to quality inspection in other industries, such as welding or gluing.

\section{Acknowledgments}

We would like to express our deep appreciation for the financial support provided by LlamaZOO Interactive Inc. and the Natural Sciences and Engineering Research Council (NSERC) Canada through the Alliance Grant ALLRP 567583 - 21. Furthermore, we would like to acknowledge the invaluable research collaboration between the University of Victoria, LlamaZOO Interactive Inc., the National Research Council of Canada (NRC), and Fives Lund LLC. Their contributions and assistance have been instrumental to the success of this project.

\bibliographystyle{ieeetr}
\bibliography{main}

\end{document}